\documentclass[10pt,journal,final,twocolumn]{IEEEtran}
\usepackage{graphicx}
\usepackage{amsmath,amssymb} 
\usepackage{color}
\usepackage{times}
\usepackage{epsfig}
\usepackage{multirow}
\usepackage{subfig}
\usepackage{tabularx}
\usepackage{graphics}
\usepackage{wrapfig}
\usepackage{url}

\allowdisplaybreaks

\input{./Definitions}
\graphicspath{{./Figures/}{../}}

\newcommand{\threevdots}{%
	\vbox{\baselineskip1ex\lineskiplimit0pt%
		\hbox{.}\hbox{.}\hbox{.}}}

\begin{document}

\title{Hand Action Detection from Ego-centric Depth Sequences with Error-correcting Hough Transform} 

\author{Chi Xu, Lakshmi Narasimhan Govindarajan\\
Bioinformatics Institute, A*STAR, Singapore\\
\and
Li Cheng,~\textit{Senior Member}\\
Bioinformatics Institute, A*STAR, Singapore, and\\
School of Computing, National University of Singapore, Singapore~\footnote{Correspondence Author, {\tt\small chengli@bii.a-star.edu.sg}.}\\
}

\maketitle

\begin{abstract}
Detecting hand actions from ego-centric depth sequences is a practically challenging problem, owing mostly to the complex and dexterous nature of hand articulations as well as non-stationary camera motion. We address this problem via a Hough transform based approach coupled with a discriminatively learned error-correcting component to tackle the well known issue of incorrect votes from the Hough transform. In this framework, local parts vote collectively for the start $\&$ end positions of each action over time. We also construct an in-house annotated dataset of 300 long videos, containing 3,177 single-action subsequences over 16 action classes collected from 26 individuals. Our system is empirically evaluated on this real-life dataset for both the action recognition and detection tasks, and is shown to produce satisfactory results. To facilitate reproduction, the new dataset and our implementation are also provided online.
\end{abstract}


\section{Introduction}

Recent development of ego-centric vision systems provide rich opportunities as well as new challenges.
Besides the well-known Google Glass~\cite{googleglass15}, more recent systems such as Metaview Spaceglasses~\cite{metaviewVR15} and Oculus Rift~\cite{oculusVR15} have started to incorporate depth cameras for ego-centric 3D vision. A commonplace shared by these ego-centric cameras is the fact that they are \emph{mobile} cameras.
Also, the interpretation of hand actions in such scenarios is known to be a critical problem~\cite{FatFarReh:iccv11,LiFatReh:iccv13,HuaEtAl:cvpr15}.
Meanwhile, facilitated by emerging commodity-level depth cameras~\cite{kinect11,softkinetic13}, noticeable progress has been made in hand pose estimation and tracking~\cite{LiKit:cvpr13,OikLouArg:cvpr14,TanEtAl:cvpr14,QiaEtAl:cvpr14,XuEtAl:ijcv15,RegSupRam:cvpr15,RegSupRam:iccv15}.
The problem of hand action detection from mobile depth sequences however remains unaddressed.

%

\begin{figure*}
\includegraphics[scale=0.47]{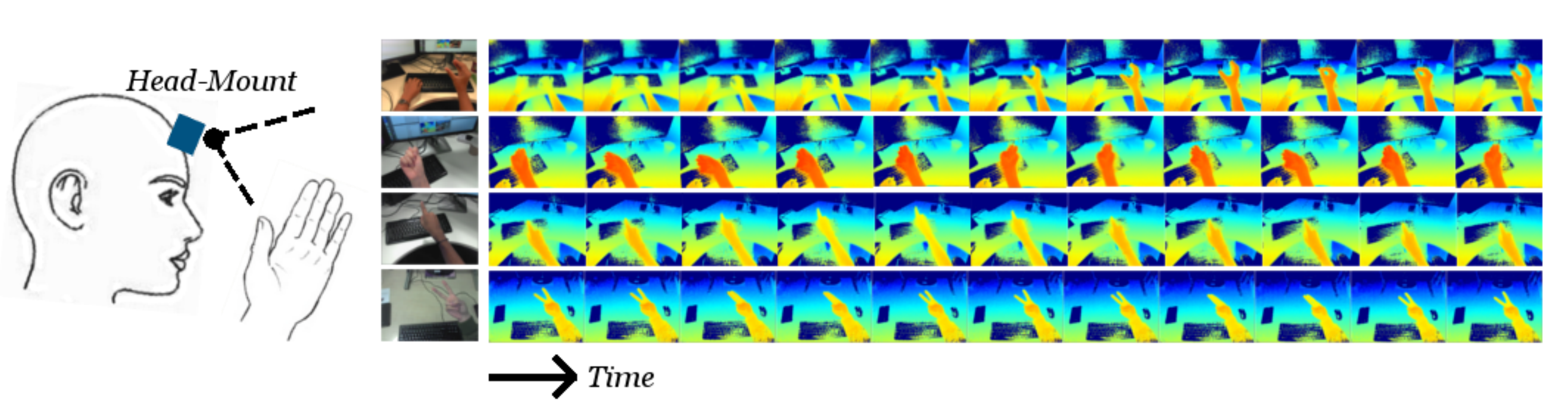}
	\caption{An illustration of the ego-centric hand action detection problem. Key frames (colorized depth images) from a few exemplar actions are shown here. The type of actions vary from coarse hand motions to fine finger motions. Action recognition/detection tasks in this scenario are challenging due to (i) Illumination artefacts, (ii) Variations across subjects in the way actions are performed, (iii) Variations in hand shapes, sizes and (iv) Non-stationary camera positions due to head motion.}
	\label{fig:dataset_sample}
\end{figure*}

As illustrated in Fig.~\ref{fig:dataset_sample}, in this paper we address this problem in the context of an ego-centric vision system.
Due to the diversity in hand shapes, sizes and variations in hand actions, it could be difficult to differentiate actions from other dynamic motions in the background. The difficulty of the problem is further compounded in the presence of a non-stationary camera as considered here. Our contribution in this paper is three-fold. (1) To our knowledge this is the first such academic effort to provide an 
effective and close to real time solution for hand action detection from mobile ego-centric depth sequences. (2) We propose a novel error-correcting mechanism to tackle the bottleneck issue of incorrect votes from Hough transform which has been shown to degrade prediction performance~\cite{RazEtAl:eccv12,WohEtAl:bmvc12,WooEtAl:ijcv13}. This follows from our observation that voting errors frequently exhibit patterns that can be exploited to gain more knowledge. (3) We make available our comprehensive, in-house annotated ego-centric hand action dataset~\footnote{This dataset and our code can be found at the dedicated project website \url{http://web.bii.a-star.edu.sg/~xuchi/handaction.htm}.} on which the proposed method is thoroughly evaluated. The error-correcting module is also examined with a series of tasks on a synthetic dataset. The empirical evaluations demonstrated that the proposed method is highly competitive and validates that our approach is able to pick out subtleties such as fine finger movements as well as coarse hand motions.


\section{Related Works}
\subsection{Action Recognition and Detection}
The problem of action recognition and detection is a classic topic in vision.
Traditionally the focus has been more on looking at full-body human activities such as ''skipping" or ''jumping"~\cite{SchLapCap:icpr04}:
For example, the problem of action detection is addressed in~\cite{PirRam:cvpr14} using context free grammars.
It has also been observed in~\cite{SchGoo:cvpr08} that very short sequences (also referred to as \emph{snippets},
usually of 5-7 frames in length) are usually sufficient to identify the action type of the entire action duration.
Single key-frames and local space-time interest point features are also utilized in~\cite{LapPer:iccv07} to detect drinking action type from realistic movie scenarios.
Yuan et al.~\cite{YuaLiuWu:cvpr09} focus on improving search efficiency, while \cite{GeoRosJit:iccv15} resorts to contexture cues and convolutional neural networks.
The work of Yao et al.~\cite{YaoGalGoo:cvpr10} is perhaps the most related, in which a probabilistic Hough forest framework is proposed for action recognition.
An interesting method is proposed in~\cite{YuKimCip:cvpr13} to use human full-body action detection to help with pose analysis.
Meanwhile, daily activity datasets of first-person color camera view has been established and studied by several groups~\cite{PirRam:cvpr12,RyoMat:cvpr13} for applications such as life-logging and tele-rehabilitation.
There are also recent works on action recognition using recurrent neural network approaches~\cite{DuWanWan:cvpr15,DonEtAl:cvpr15}.
Very recently, more research efforts have focused on action recognition from depth sequences~\cite{YeEtAl:bookchapter13}:
For instance, an action-let ensemble model is proposed in~\cite{WanEtAl:pami14} to characterize individual action classes and
intra-class variances; The problem of multi-label action detection problem is considered in \cite{WeiEtAl:iccv13} with a structural SVM model.
The work of~\cite{MogEtAl:cvprwshp14} is among the few efforts to further explore head-mount RGB-D camera for action recognition.
Nevertheless, hand action detection still lacks thorough investigation, especially for depth cameras in the context of mobile ego-centric vision.
%

Related works on \emph{hand} action recognition and detection are relatively scarce.
Among the early works on hand gesture recognition, Lee and Kim~\cite{LeeKim:tmapi99} study the usage of a hidden Markov model (HMM) based on hand tracking trajectories.
The emergence of consumer depth cameras significantly revolutionized the landscape of this field,
where upper or full body skeleton estimation~\cite{ShoEtAl:cacm13} is shown to be a powerful feature for the related problem of sign language recognition~\cite{lang2012sign}.
Recently, multi-sensor hand gesture recognition systems~\cite{molchanov2015multi,ohn2014hand} are proposed in a car driving environment with stationary rear cameras aimed at the driver.
We note in passing that in-car cameras are usually stationary while in this paper the ego-centric vision refers to the general case of non-stationary cameras, and in particular we look at head-mount cameras.
Therefore, adjacent image frames are not aligned any more and usual techniques such as background subtraction are not applicable.
Furthermore, the Kinect human body skeleton estimation becomes inapplicable as the camera is head-mounted.

\subsection {Hough Transform}
As probably one of the most widely used computer vision techniques, the Hough transform is first introduced by~\cite{Hou:ICHEAI59},
initially as a line extraction method.
It is subsequently reformulated by~\cite{DudHar:CACM72} to its current form and extended to detect circles.
Furthermore, Ballard~\cite{Bal:PR81} develops the Generalized Hough transform to detect arbitrary shapes.
In a \textit{typical} Hough transform procedure, a set of local parts is captured to sufficiently represent the object of interest.
Then, each of the local parts vote for a particular position~\cite{Bal:PR81}.
Finally in the voting space, computed as \emph{score function}s of the vote counts over all parts, an object is detected by mode-seeking the locations receiving the most significant score.

An important and more recent development due to~\cite{LeiLeoSch:eccv04whp} is its extension to a probabilistic formulation to incorporate a popular part-based representation,
the visual Bag-of-Words (BoW). This is commonly referred to as the {\em Implicit Shape Model} (ISM), as follows.
Formally, consider an object of interest as $x$, and its label as $y \in \mathcal{Y}$, where $\mathcal{Y}$ is a set of feasible object categories.
The position of the object in voting space is characterized by e.g. its nominal central location and scale in images or videos, and is collectively denoted as $\Lambda$.
In our context, the object $x$ is captured by a number of (say $I$) local parts, $x = \{e_i\}_{i=1}^I$, and the number may vary from one object to another.
Denote by $f_i$ the feature descriptor of a local part which is observed at a \textit{relative}
location~\footnote{The relative location is usually measured with respect to the object's nominal center.} $l_i$.
As a result, $e_i := \{f_i, l_i\}$.
We consider a visual Bag-of-Words representation, and denote by $c_k$ the $k$-th codebook entry of the weight vector for a local part $e_i$.
The score function now becomes
\begin{align}
	\nonumber
	S_{\Lambda}^{\textrm{ISM}}(y) &:= \sum_{k,i} p(\Lambda, y, \{c_k\}, \{f_i, l_i\} ) \\
	\nonumber
	&= \sum_{k,i} p(\Lambda|y, c_k, l_i) p(y|c_k, l_i) p(c_k|f_i) p(f_i) p(l_i) \\
	&\propto \sum_i  \sum_k p(y|c_k, l_i) p(\Lambda|y, c_k, l_i) p(c_k|f_i),
	\label{eq:score_func_ism}
\end{align}
which recovers Eq.(5) of~\cite{LeiLeoSch:eccv04whp}.
The voting space is thus obtained by computing the score function over every position $\Lambda$.
The seminal work of ISM~\cite{LeiLeoSch:eccv04whp} has greatly influenced the recent development of Hough transform-like methods (e.g.~\cite{MajMal:cvpr09,GalEtAl:pami11}),
where the main emphases are on improving voting and integration with learning methods.
Specifically the large-margin formulation of Hough Transform~\cite{MajMal:cvpr09} is closely related.
%
Moreover, the generality of Hough transform on how each of the local parts could vote enables myriad part-to-vote conversion strategies.
Instead of the BoW strategy of~\cite{LeiLeoSch:eccv04whp,MajMal:cvpr09},
Hough forests~\cite{GalEtAl:pami11} are proposed which can be regarded as generating discriminatively learned codebooks.

There are also related works examining Hough vote consistency in the context of 2D object detection and recognition.
A latent Hough transform model is studied in~\cite{RazEtAl:eccv12} to enforce the consistency of votes.
In \cite{yarlagadda2010voting}, the grouping, correspondences, and transformation of 2D parts are initialized by a bottom-up grouping strategy,
which are then iteratively optimized till convergence.
In \cite{barinova2012detection}, the hypothesis and its correspondences to the 2D parts are updated by greedy inference in a probabilistic framework.
Our work is significantly different from these efforts:
(1) Instead of exploring only the consistency of correct votes, we also seek to rectify the impact of incorrect votes, such that both correct and incorrect votes can contribute to improve the performance.
(2) Rather than iteratively optimize complex objectives, with many unknown variables~\cite{yarlagadda2010voting,barinova2012detection}, that are often computationally expensive and prone to local minima, our algorithm is simple and efficient as the process involves only linear operations.
(3) The above mentioned algorithms are designed for 2D object detection, which cannot be directly applied in the temporal context considered in this paper.

\subsection{Related work on Error Correcting Output Codes}
The concept of error correcting or error control has been long established in information theory and coding theory communities~\cite{HufPle:book03},
which has been recently employed for multiclass Classification~\cite{DieBak:jair95}.
It is shown in~\cite{shreyas2009} that there is an interesting analogy between the standard Hough transform method and error correction codes in the context of curve fitting.

\begin{figure}
	\centering\includegraphics[width=0.9\columnwidth]{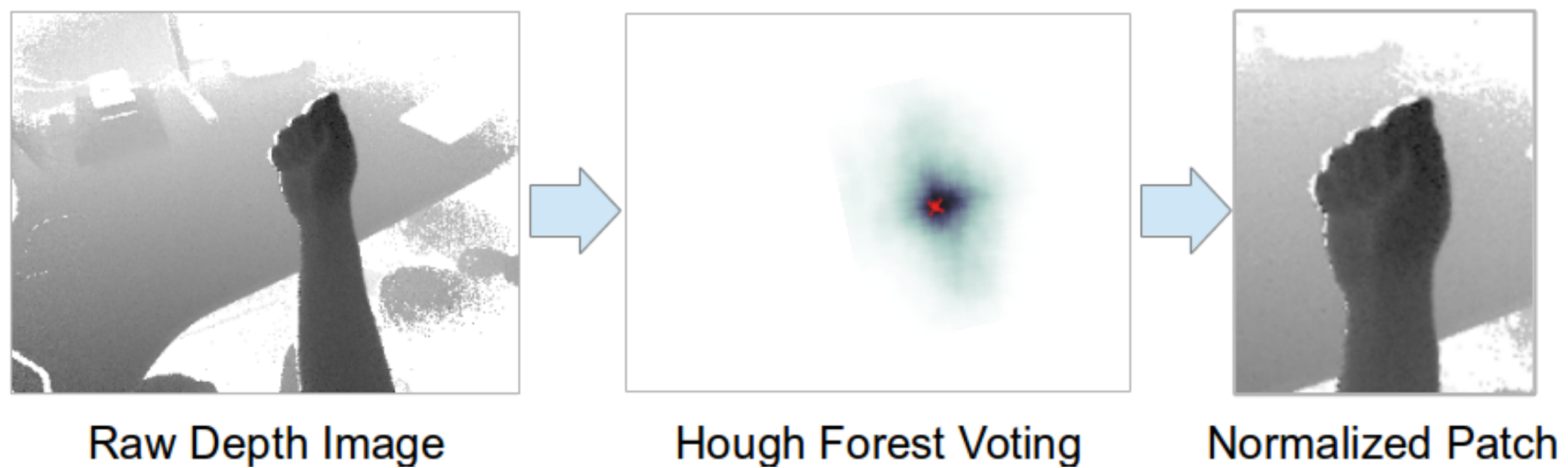}
	\caption{An illustration of the preprocessing step on hand normalization.}
	\label{fig:normalize_patch}
\end{figure}
\section{Our Approach}

Inspired by \cite{SchGoo:cvpr08}, we consider the employment of snippets as our basic building blocks, where a temporal sliding window is used to densely extract snippets from video clips.
Each snippet thus corresponds to such a temporal window and is subsequently used to place a vote for both the action type and its start $\&$ end positions under the Hough transform framework.
This dense voting strategy nevertheless leads to many uninformative and even incorrect local votes where either
the action type or its start/end locations could be wrong.
In what follows we propose an error correcting map to explicitly characterize these local votes, where
the key assumption here is that for a particular action type, the patterns of accumulated local votes are relatively stable.
The patterns refer to spatial and categorical distributional information of the collection of local votes
obtained from the snippets in training set, which include the \emph{correct} local votes as well as the \emph{incorrect} votes.

\begin{figure}
	\centering
	\includegraphics[width=0.95\columnwidth]{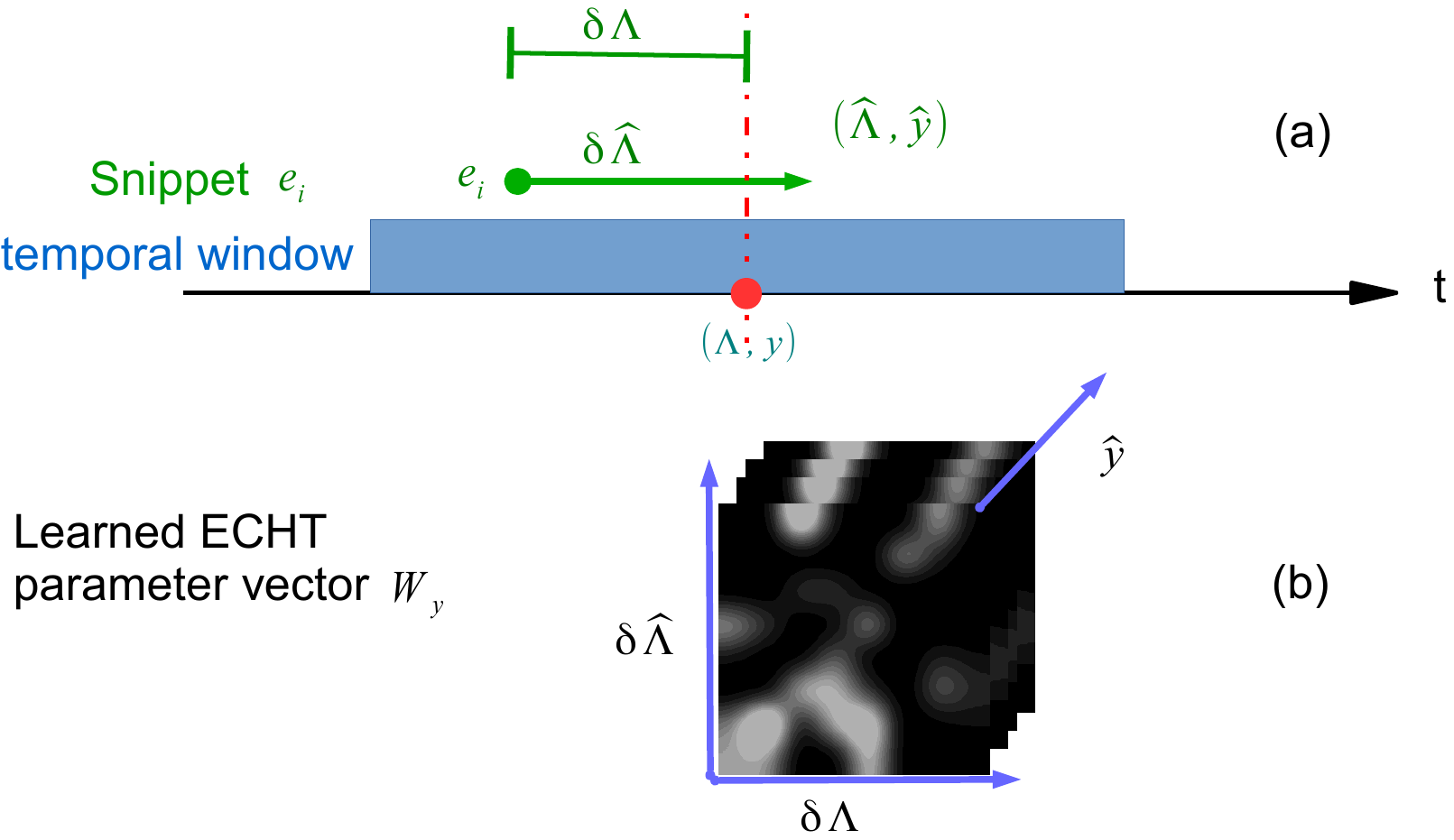}
	\caption{An illustration of the proposed error correcting Hough transform. (a) Each snippet $e_i$ votes on the action type $\hat{y}_i$ and its center position $\hat{\Lambda}_i$.
			$\delta \hat{\Lambda}_i$ denotes the amount of deviation,
			while $\delta \Lambda_i$ being the difference between the snippet position and the true center position. Note the center position is nominally shown here only for illustration purpose.
			In practice, the start or end positions are used instead. (b) The learned $W_y$ vector of an action type $y$, which is re-organized into the error correcting parameter cubic space. }
	\label{fig:echt}
\end{figure}

\begin{figure*}
	\centering\includegraphics[width=0.99\linewidth]{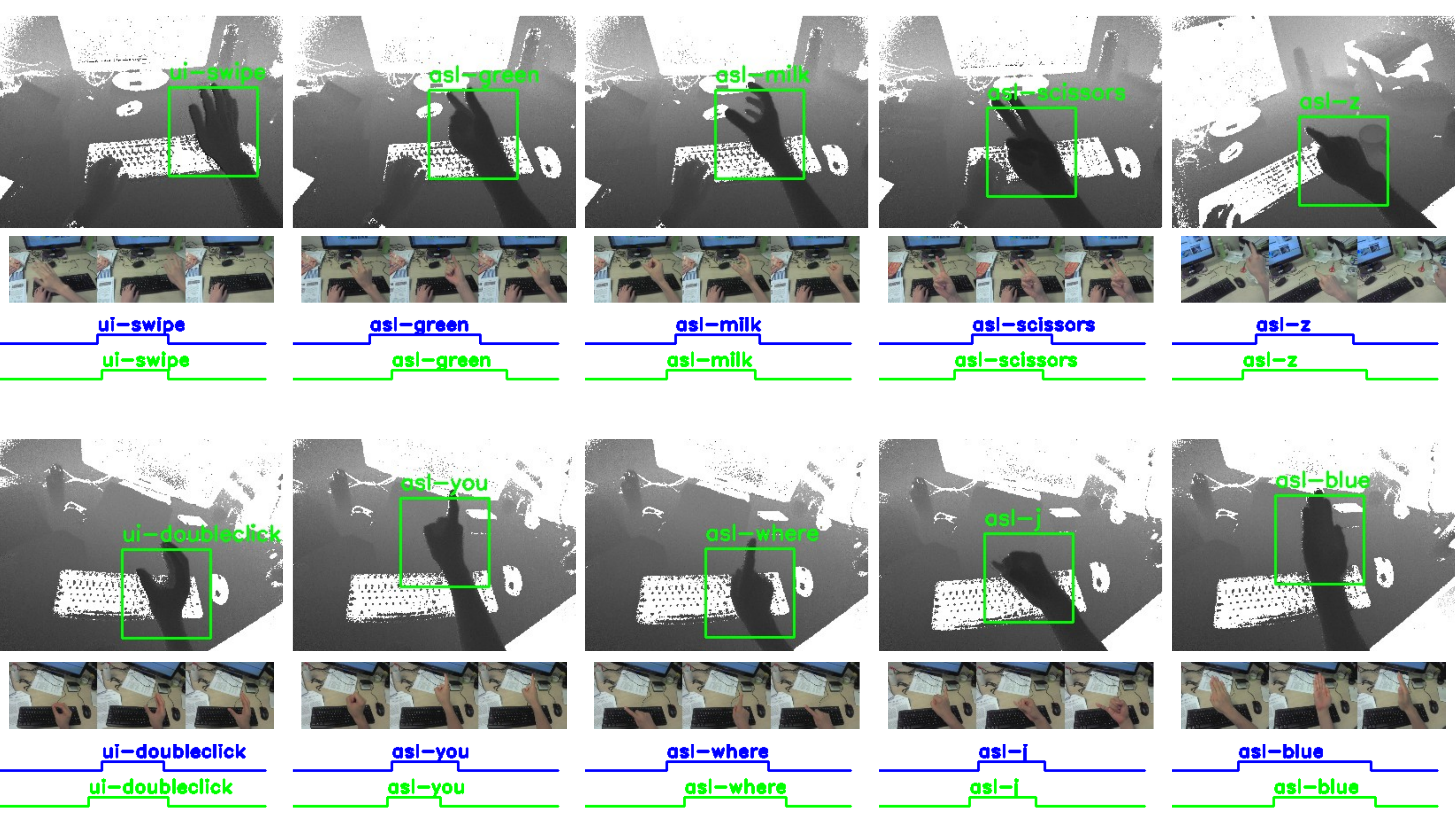}
	\caption{Exemplar hand action detection results. In this and next Figures, blue line denotes the ground-truths,
			while green line is for the correct detection, and red line for the incorrect detection.}
	\label{fig:visual}
\end{figure*}

\paragraph{A Preprocessing Step: Hand Normalization}
To facilitate the follow-up feature generation, as a preprocessing step we 
normalize the hand position, in-plane orientation, and size.
Assume that a hand is present in each depth image.
During this step the hand location, orientation, and size are estimated using Hough forest~\cite{XuEtAl:ijcv15}. 
This is achieved by jointly voting for the 3D location, in-plane orientation, and size of the hand on each pixel, then the local maxima in the accumulated voting space is located by mean-shift~\cite{XuEtAl:ijcv15}. 
Based on these triplet information, the hand image patches are further normalized such that in what follows we would work with a canonical upright hand
with a fixed patch size as exemplified in Fig.~\ref{fig:normalize_patch}.
In this paper the patch size is fixed to $120 \times 160$ pixels.


\paragraph{Local Votes from Snippets}

After the aforementioned preprocessing step, a sliding window of length $w_s$ is used to densely extract (with step size 1) snippets from the temporal zones in video clips where hand actions take place.
Each snippet thus corresponds to a temporal subsequence of fixed length $w_s$, and at each frame, it contains the normalized hand patch as well as the estimated hand location and orientation.
At test run, a snippet is subsequently used to place a vote for the \emph{action type} as well as its \emph{start} and \emph{end} positions.

To simplify the matter, for the target position to be voted, the center position of current sliding window is nominally used here only for illustration purpose as in Fig.~\ref{fig:echt}(a).
In practice, the start or end positions are used instead. In other words, the nominal center position is replaced in practice by either the start or the end positions separately.
Now, consider a relatively larger sliding temporal window of length $w_t > w_s$ that contains multiple snippets.
We further denote its nominal center location $\Lambda$, and assume that this temporal window is overlapped with one of the action subsequences with action type $y$.
This sliding window will be the corner-stone of our paper in producing the Hough voting score.
As illustrated in Fig.~\ref{fig:echt}(a), given a snippet $e_i$ located at $l_i$ in such a temporal window,
let $\delta \hat{\Lambda}$ denote the quantized temporal deviation from the snippet location to a position $\hat{\Lambda}$ it might vote for, and $\hat{y}$ be a possible action type.
Denote $\delta \Lambda$ the quantized deviation from the snippet location to the center position of current sliding window.
The quantization of $\delta \hat{\Lambda}$ and $\delta \Lambda$ is necessary here since the temporal frames are already quantized.
Now we would like to learn the probability distribution of temporal offsets $\delta \Lambda$, $\delta \hat{\Lambda}$ and class label $ \hat{y}$, which can be factorized as
\begin{align}
	p(\delta \Lambda, \delta \hat{\Lambda}, \hat{y} | e_i) = p(\delta \Lambda | e_i) p(\delta \hat{\Lambda}| \hat{y}, e_i) p(\hat{y}|e_i).
	\label{eq:cond_prob}
\end{align}
Two random forests are trained for this purpose: The first forest, a classification random forest $\mathcal{T}_c$, models the classification
probability $p(\hat{y}|e_i)$, while the second one is a conditional regression forest $\mathcal{T}_{r}$ that represents the conditional probability $ p(\delta \hat{\Lambda}| \hat{y}, e_i)$.
The term $p(\delta \Lambda | e_i)$ explicitly displays the quantization process: $\delta \Lambda$ is a random variable indexing over the bins after quantization.
Then $p(\delta \Lambda | e_i) = 1$ for one bin where the deviation of location of $e_i$ to $\Lambda$ falls exactly under it,
and $p(\delta \Lambda | e_i) = 0$ when $\delta \Lambda$ refers to any other bins.
For both forests, two sets of features are used for binary tests in the split nodes:
The first is the commonly-used set of features that measures the absolute difference of two spatio-temporal 3D offsets~\cite{TanEtAl:cvpr14} in the normalized hand patches.
It is complemented by the second set of features which considers the 6D parameters of estimated hand location and orientation from hand localization.
In addition, the standard Shannon entropy of multivariate Gaussian distribution~\cite{XuEtAl:ijcv15} is used to compute information gains at the split nodes.

Votes from these local snippets however produce a fair amount of local errors.
These voting errors could be categorized into two types: large temporal deviation (the start $\&$ end positions) and inter-class confusion.
The large temporal deviation arises from various reasons, including
\begin{itemize}
  \item Temporal rate variation:  The scale and speed of actions may vary notably in practice across subjects and at different times;
  \item Repetitive pattern: For example, in action ``asl-blue'' that will be formally introduced later, the hand is flipped twice, and it is difficult to predict whether it should be the first flip or the second, when
	one only focuses on a short action snippet at a time;
  \item Manual labeling inconsistency: The annotation of start and end positions may be inconsistent in training data.
\end{itemize}
%
Meantime, inter-class confusion refers to scenarios when a snippet is mistakenly categorized into an incorrect action type.
For example, since a snippet extracted from action ``asl-milk'' could be similar to that from action ``ui-doubleclick'', they may be confused when placing the local vote.
An important observation here is that both temporal deviations and inter-class confusions often exhibit specific patterns that could be exploited potentially.
These observations lead us to propose in what follows a novel mechanism to cope with and even benefit from these local errors.

\paragraph{Our Error Correcting Hough Transform (ECHT)}
The central piece of Hough transform lies in the score function defined for the voting space, as has been elaborated in the previous section.
In this paper, we consider a {\em linear additive} score function of an object with label $y$ at position $\Lambda$ as,
\begin{align}
	S_{\Lambda}(y) := \sum_i s_i
	= W_y^T \, \Phi_{\Lambda}(x).
	\label{eq:score_func_linearform}
\end{align}
The score function is additive in term of local votes with each being defined as $s_i := W_y^T \, \phi_{\Lambda}(e_i)$.
Therefore the features is decomposed into local snippet-based features $\Phi_{\Lambda}(x) := \sum_i \phi_{\Lambda}(e_i)$.
Consider for example the large-margin formulation of Hough Transform in~\cite{MajMal:cvpr09}, which amounts to being
a relaxation of \eqref{eq:score_func_ism} to non-probabilistic function forms:
by reducing $p(y|c_k, l_i)$ to $p(y|c_k)$,
and by denoting the weight vector as $w_y := (\cdots, p(y|c_k), \cdots)$, the linear form of \eqref{eq:score_func_linearform} is obtained,
with $\Phi_{\Lambda}(x) = \sum_i \phi_{\Lambda}(f_i, l_i) =  \sum_i p(\Lambda|y, c_k, l_i) p(c_k|f_i)$.
This is exactly the large-margin formulation of ISM described in Eq.(12) of~\cite{MajMal:cvpr09}.

In our context, since the local votes from snippets contain noticeable amount of errors, it is necessary to consider an error-control mechanism.
This motivates us to consider, instead of the BoW approach as studied in e.g.~\cite{LeiLeoSch:eccv04whp,MajMal:cvpr09},
a new linear map that explicitly characterizing errors from local votes:

\begin{align}
	\nonumber
	S_{\Lambda}^{\textrm{EC}}(y) 
	&= \sum_{i} \sum_{\delta\Lambda, \delta\hat{\Lambda}, \hat{y}} p \big(\Lambda, y, \{\delta\Lambda, \delta\hat{\Lambda}, \hat{y}\}, \{e_i\}\big) \\
	\nonumber
	&\propto \sum_{\delta\Lambda, \delta\hat{\Lambda}, \hat{y}}  p\big(\Lambda, y | \delta \Lambda, \delta \hat{\Lambda}, \hat{y} \big) \, \sum_i  p\big(\delta \Lambda, \delta \hat{\Lambda}, \hat{y} | e_i\big) \\
	&= W_y^T \, \Phi_{\Lambda} (x).
	\label{eq:score_func_our}
\end{align}
As illustrated in Fig.~\ref{fig:echt}(b), given an action type $y$,
$W_y = \begin{pmatrix}
\threevdots\\
w^{(y)}_{\delta\Lambda, \delta\hat{\Lambda}, \hat{y}}\\
\threevdots
\end{pmatrix}
$

\begin{figure}
	\centering
	\includegraphics[width=0.41\columnwidth]{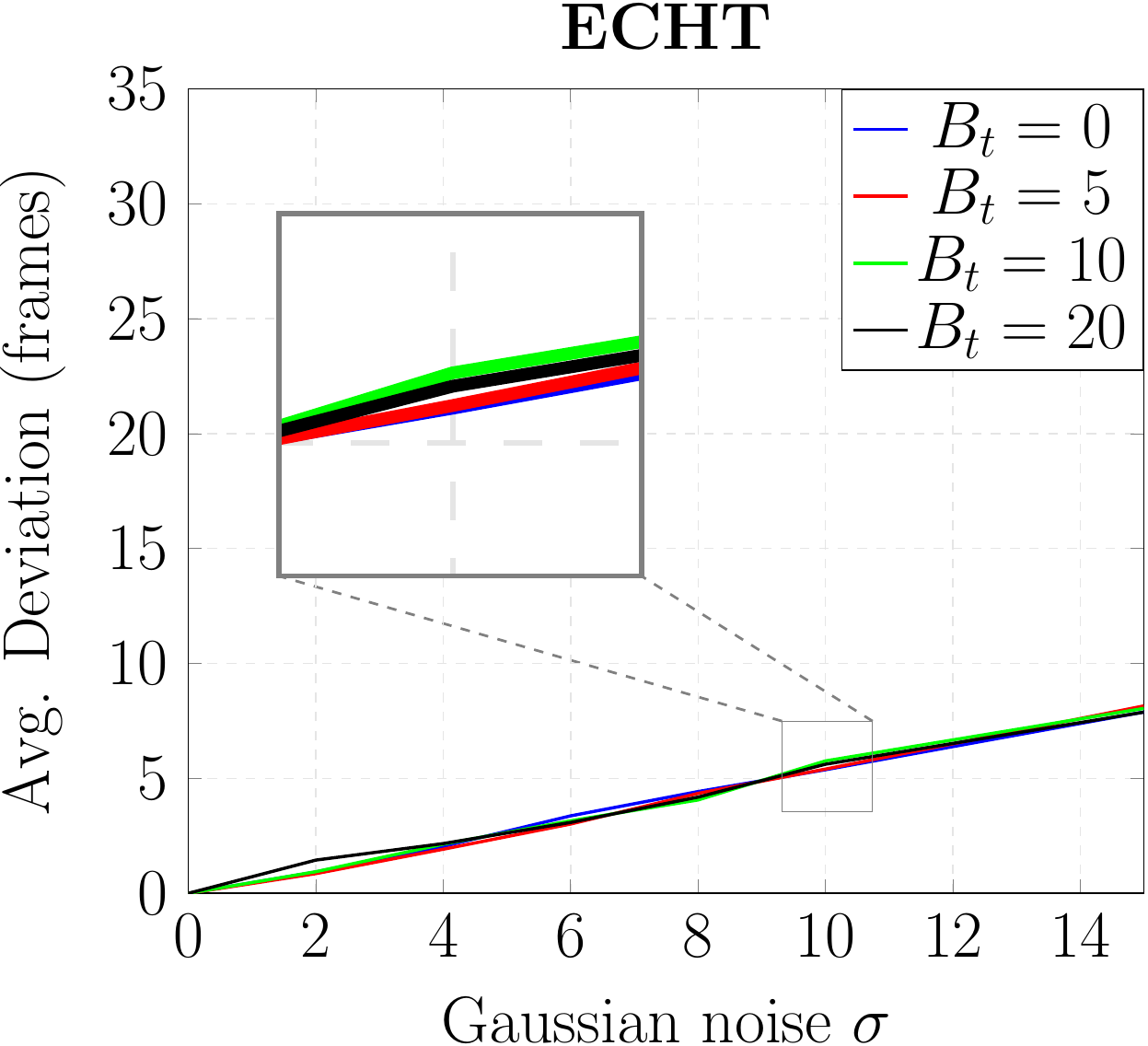}
\hspace{2mm}
	\includegraphics[width=0.41\columnwidth]{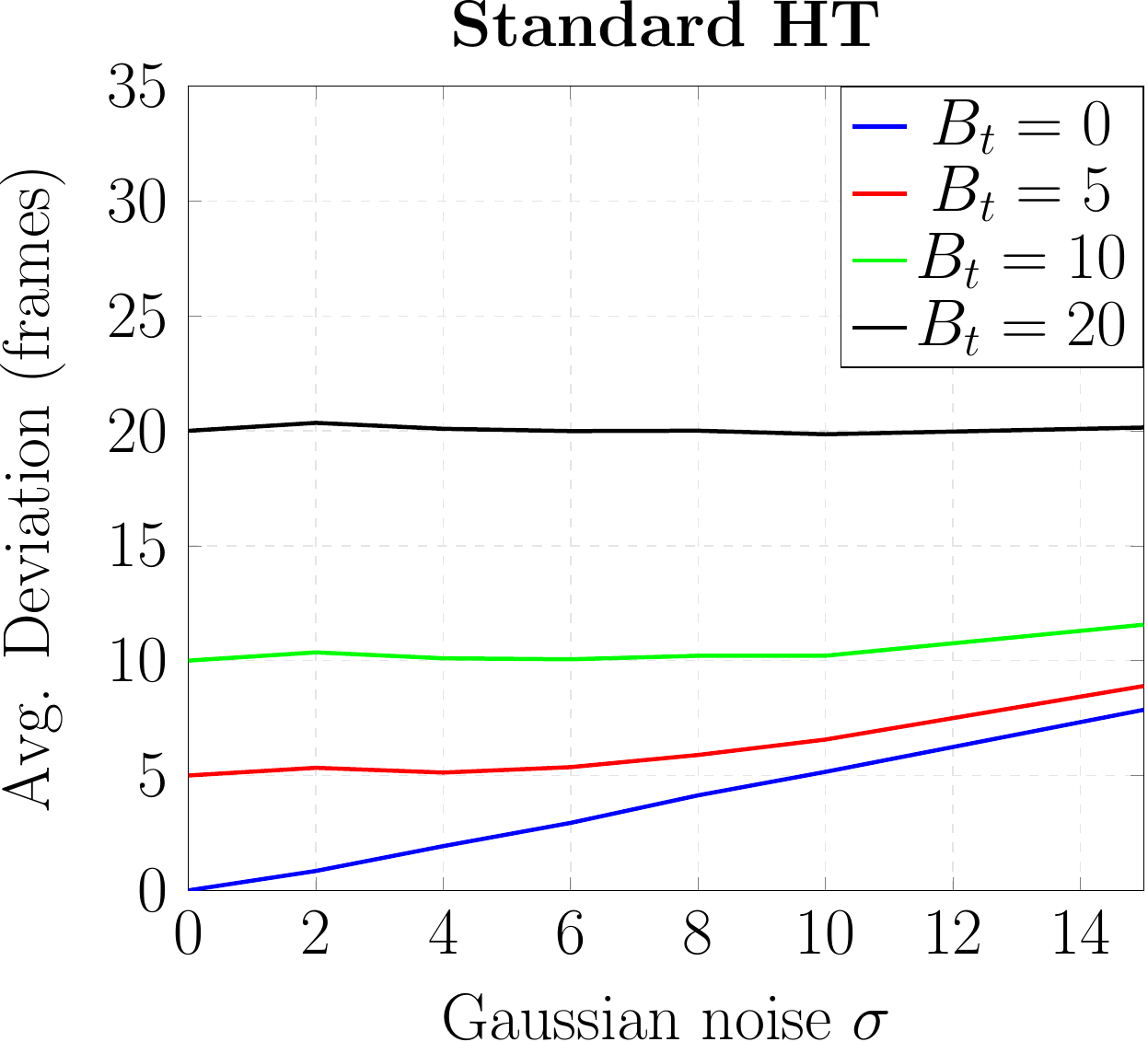}
%
	\caption{A systematic temporal error analysis using the synthetic dataset. $x$-axis denotes the temporal additive Gaussian noise with zero mean and standard deviation $\sigma$, while $y$-axis presents the average temporal deviations from the ground-truths. }
	\label{fig:syntheticExpBias}
\end{figure}

denotes a column vector obtained by concatenating the parameters over the range of three-dimensional error correcting space of $\delta \Lambda$, $\delta \hat{\Lambda}$ and $\hat{y}$,
with each element $ p\big(\Lambda, y | \delta \Lambda, \delta \hat{\Lambda}, \hat{y} \big)$ denoting a particular parameter.
In a similar way, each element of $\Phi_{\Lambda} (x)$ is $\sum_i p\big(\delta \Lambda, \delta \hat{\Lambda}, \hat{y} | e_i\big)$,
which encodes the local vote obtained from the snippet $e_i$ using random forests as in Eq.~\eqref{eq:cond_prob}.
$p(e_i)$ is uniformly distributed and is thus ignored as being a constant factor.

\paragraph{Training $\&$ Testing Phases of ECHT}

Since the Hough voting space is characterized as a linear map $S_{\Lambda}^{\textrm{EC}}(y)= W_y^T \, \Phi_{\Lambda} (x)$,
we can learn the parameter vector $W_y$ during the training phase as follows.
For each of the action subsequences in video clips, we randomly sample subsequences around it.
For each such subsequence $x_i$ of action type $y_i$ and start/end positions $\Lambda_i$,
its ground-truth voting score $S_{\Lambda_i}(y_i)$ is defined as the intersection over union of the sampled subsequence and the action subsequence.
The training objective amounts to estimating the parameter $W_y$ that minimizing the discrepancy between the ground-truth and the computed scores over all training subsequences,
plus a regularization term of $W_y$:
\begin{align}
	W_y^*=\arg\min_{W_y} \frac{1}{2} \|W_y\|^2 + c \sum_{i} \mathrm{d}
	\Big(
	S_{\Lambda_i}(y_i) - W_y^T \, \Phi_{\Lambda} (x_i)
	\Big),
	\label{eq:kernel_objFunc}
\end{align}
where $\|\cdot\|$ denotes the vector norm,
$c$ a trade-off constant, $i$ indexes over all the training subsequences, and $\Phi_{\Lambda} (x_i)$ considers all the snippets within the current subsequence $x_i$.
In this paper, we consider the $\epsilon$-insensitive loss~\cite{ChaLin:atist11} for $\mathrm{d}(\cdot)$
and the problem is solved by linear support vector regression (SVR)~\cite{ChaLin:atist11}.

%

At test run, given a test example consisting of a set of snippets $x = \{e_i\}$,
the action detection problem boils down to finding those $S_{\Lambda}^{\textrm{EC}}(y) \geq \delta_{\mathrm{tst}}$ with a threshold $\delta_{\mathrm{tst}}$
and with the help of non-maximal suppression.
%

\section{Experiments}

An in-house dataset has been constructed for hand action detection as illustrated in Fig.~\ref{fig:dataset_sample},
where the videos are collected from a head-mount depth camera (i.e. time-of-flight depth camera Softkinetic DS325).
The spatial resolution of this depth sensor is $320\times240$ pixels, while the horizontal and vertical field-of-view are $74^{\circ}$ and $58^{\circ}$, respectively.
The video clips are acquired at a frame rate of 15 frames per second (FPS).
We consider 16 hand action classes, with ten classes from ASL (American sign language) and the rest six from UI (user interface applications), as follows:
\textit{asl-bathroom, asl-blue, asl-green, asl-j, asl-milk, asl-scissors, asl-where, asl-yellow, asl-you, asl-z, ui-circle, ui-click, ui-doubleclick, ui-keyTap, ui-screenTap,
	ui-swipe}.
During data acquisition, the distance of hand to camera varies from 210mm to 600mm, with the average distance of 415mm.

The following methods are considered during empirical evaluations:
\begin{itemize}
  \item ECHT: The proposed full-fledged approach with an error correcting map for both temporal error and inter-class confusion.
  \item ECHT-T: A degraded variant of ECHT with only temporal error correction.
  \item ECHT-C: A degraded variant of ECHT with only inter-class confusion correction.
  \item Standard HT: Standard Hough forest method using a Dirac function for class prediction,
	and a Gaussian distribution smoothing function over the estimations of temporal start/end positions,
	which can be regarded as a adaptation of the state-of-the-art method~\cite{YaoGalGoo:cvpr10}) in our problem.
  \item HMM1 \& HMM2: Two variants of the standard HMM for action recognition tasks following e.g.~\cite{LeeKim:tmapi99} to train a HMM for each action class.
	For HMM1 we use the normalized 3D hand movement feature and HoG features from normalized hand patches, while for HMM2 we use only the HoG feature.
\end{itemize}
%
%
%

Throughout our experiments, for both Hough forests $\mathcal{T}_c$ and $\mathcal{T}_{r}$, the number of trees is set to 20, and the tree depth is set to 20 by default.
$\hat{\Lambda} \in \mathbb{R}$ is used to refer to either the start or end positions of current action subsequence, as these two boundary positions are voted independently.
The error correcting cubic space of $\delta \Lambda$, $\delta \hat{\Lambda}$ and $\hat{y}$ are quantized into $8 \times 8 \times 16$ sub-cubes.
$\epsilon=0.01$ and $c=1$ for the SVR problem~\cite{ChaLin:atist11}.

In term of evaluation criteria,
a correct prediction is defined as an image subsequence with intersection-over-union ratio greater than 0.5 when comparing to the ground-truth action subsequence,
and with correctly predicted action label.
This naturally leads to the consideration of using precision, recall, as well as F1 score as the performance evaluation criteria.
Following standard definition, F1 score is a harmonic mean of precision and recall, which is defined as $F1 = 2 \cdot \frac{precision \cdot recall}{precision+recall}$.

\begin{figure}
	\centering
	\includegraphics[width=0.41\columnwidth]{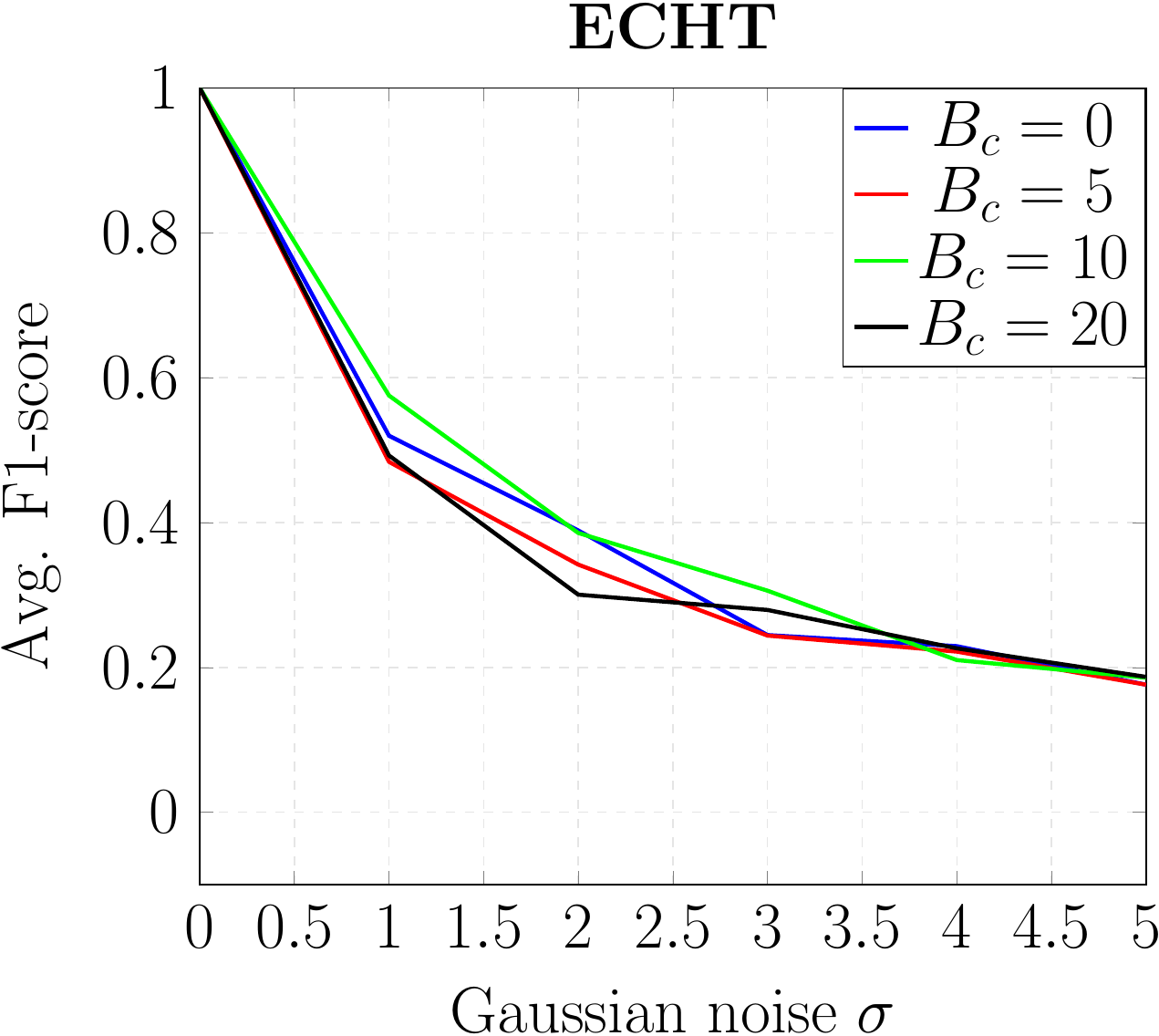}
\hspace{2mm}
	\includegraphics[width=0.41\columnwidth]{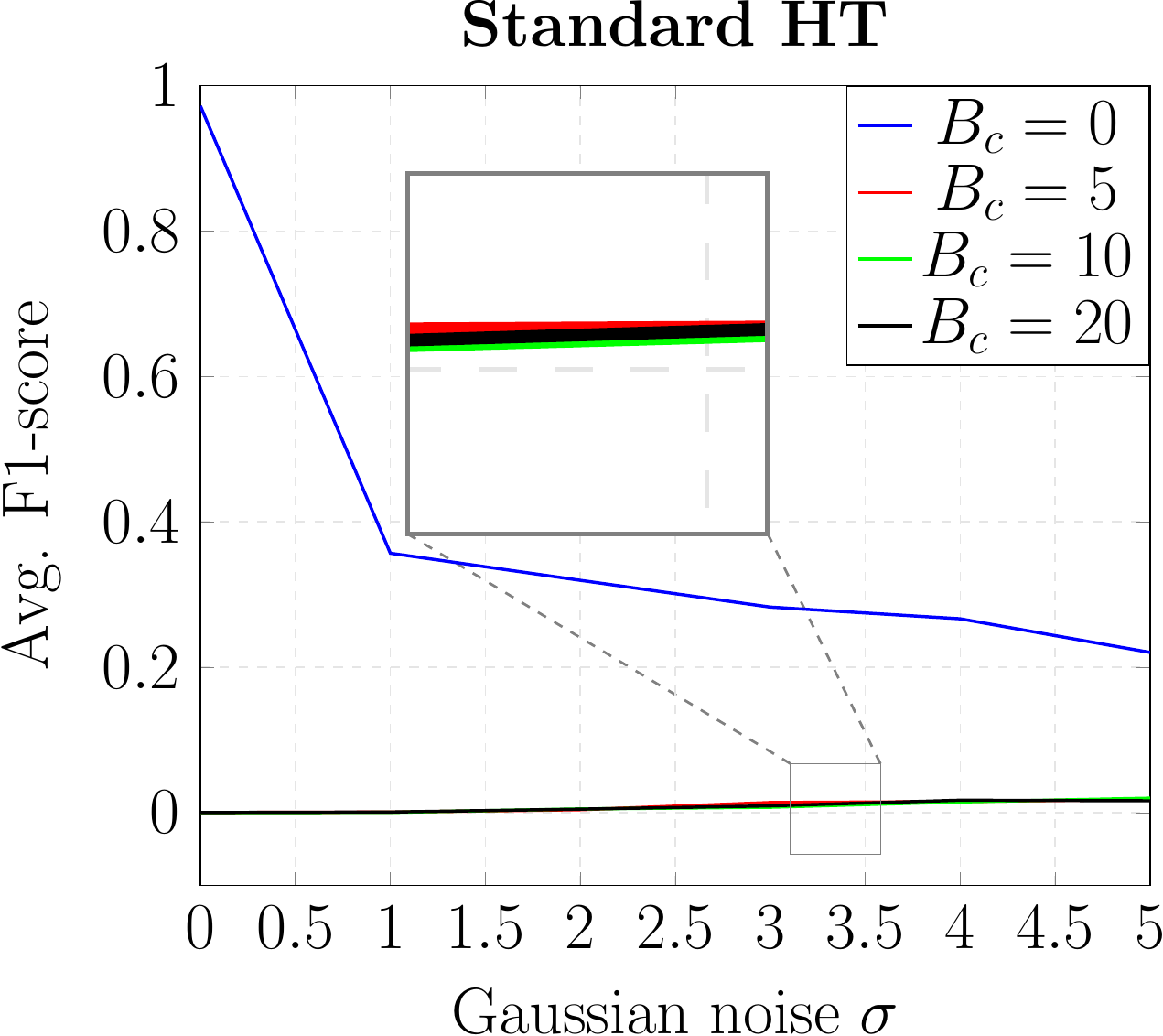}
%
	\caption{A systematic inter-class error analysis using the synthetic dataset. $x$-axis denotes the additive Gaussian noise with zero mean and standard deviation $\sigma$,
			while $y$-axis presents the average F1 score. }
	\label{fig:syntheticExpClass}
\end{figure}

\subsection{Synthetic Experiments}
To facilitate in-depth analysis of the proposed ECHT approach under a controlled environment, we carry out a series of experiments on a synthetic dataset simulating simplified scenarios.
During these experiments, a training set of about 2,500 such short temporal sequences with 16 action classes and test set of 5 long sequences with assorted actions are generated.
In practice, errors in the local votes could be due to a systematic bias (i.e. a simple version of it is all the votes are added up by a fixed constant),
random (i.e. addition Gaussian noise) or a combination of both.
Large magnitude random errors would be hard for any mechanism to correct over.
On the other hand, we hypothesize that errors of systematically biased nature can be fully accounted for and corrected for our ECHT approach,
while this type of errors remain non-correctable for standard HT.
To demonstrate this on temporal and inter-class errors we performed two sets of experiments.

\paragraph{Experiments on Temporal Error Analysis}
Here each local vote for the start point of its action sequence is purposely left-shifted by a value equal to $B_t$ along with an additional Gaussian noise with standard deviation $\sigma$.
Over a single train-test pass, the value of $B_t$ is kept constant.
There are no errors introduced in the votes for the action type.
Figures~\ref{fig:syntheticExpBias}(a) and (b) show the average deviation in the predicted start location of actions from the ground truth for various $\sigma$ and $B_t$.
For the trivial case of $\sigma=0$ and $B_t=0$, both ECHT and the standard HT deliver exact predictions.
For $\sigma=0$ but with increasing $B_t$, there is a corresponding decrease in performance of the standard HT, which relies purely on the local votes.
In contrast, ECHT manages to account for this systematic shift and produce exact predictions.
With increasing $\sigma$ there is a degradation in the performance of both methods.
which is consistent with the fact that random noise is hard to control for.
Note ECHT-T and ECHT-C are not shown and the former one performs exactly the same as ECHT while the latter one being the same as the standard HT.
Besides, to produce this and the next figures, ten experimental repeats are performed independently for each parameter set.

\begin{figure}
	\centering\includegraphics[width=0.95\columnwidth]{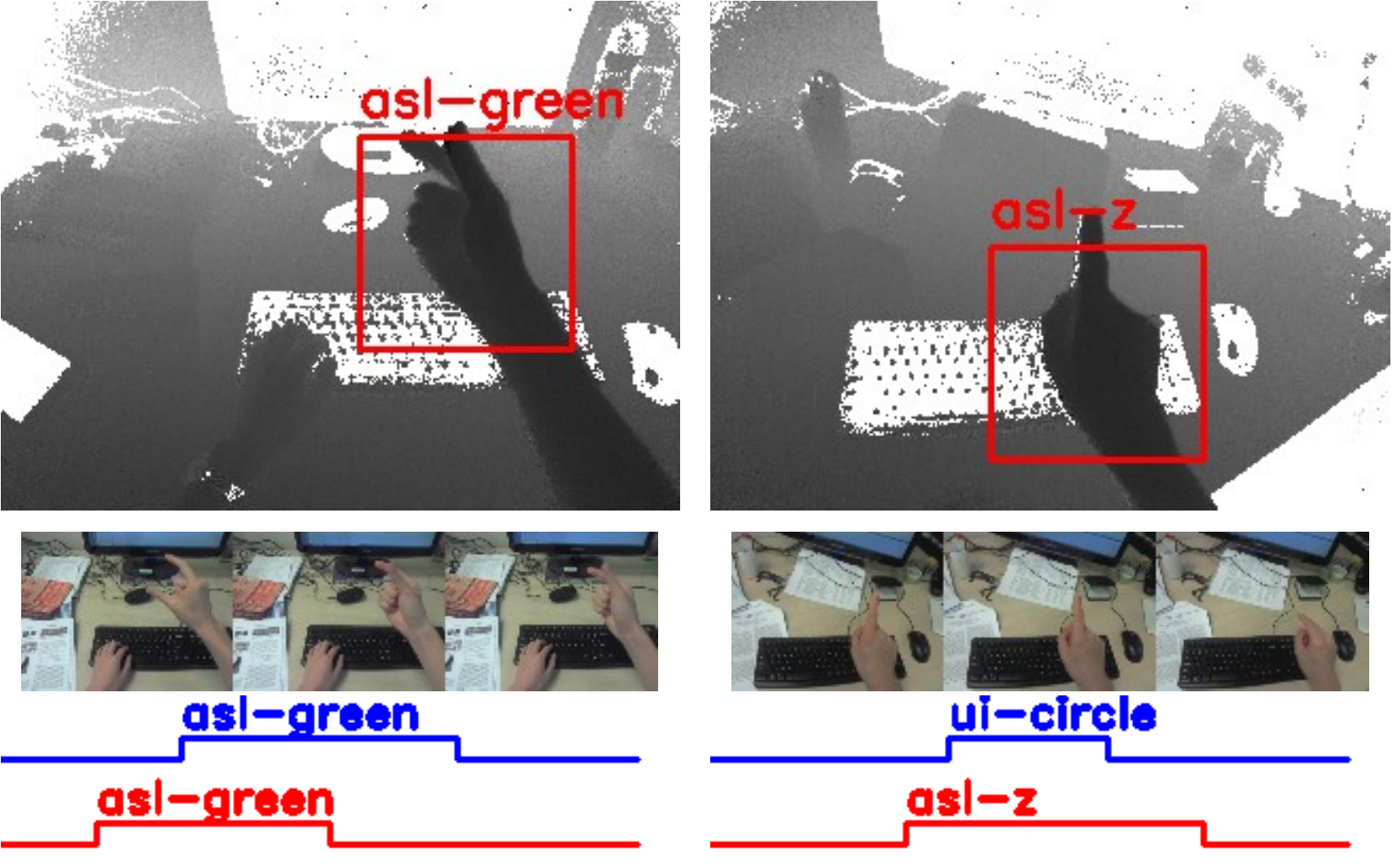}
	\caption{Incorrect detections. Left: The temporal overlap is insufficient. Right: Action class is predicted wrongly. }
	\label{fig:falseDetExmples}
\end{figure}

\paragraph{Experiments on Inter-class Error Analysis}
In these set of experiments, the local votes for the start and end positions of actions are untouched.
Instead, all the votes for action type are perturbed. Initially each class vote is a binary probability distribution rolling over all the action types.
First we cyclically rotate this distribution $B_c$ times. Then a Gaussian noise $\sigma$ is added,
and the vote is assigned a new class id sampled according to this modified distribution.
Figures~\ref{fig:syntheticExpClass}(a) and (b) present the average F1-score for various levels of $B_c$ and $\sigma$.
For $\sigma=0$, ECHT produces 100\% accurate predictions for all values of $B_c$, as the error correcting map is able to control for systematic swap in the class ids.
On the other hand, for standard HT the performance drops to 0\% for non-zero values of $B_c$ ($\sigma=0$).
Complete reliance on the local votes explains this result as for $B_c>0$ every vote predicts a class id other than the true class.
For increasing values of $\sigma$, the performances of both methods degrade.

\begin{figure}
	\centering
	\includegraphics[width=0.95\columnwidth]{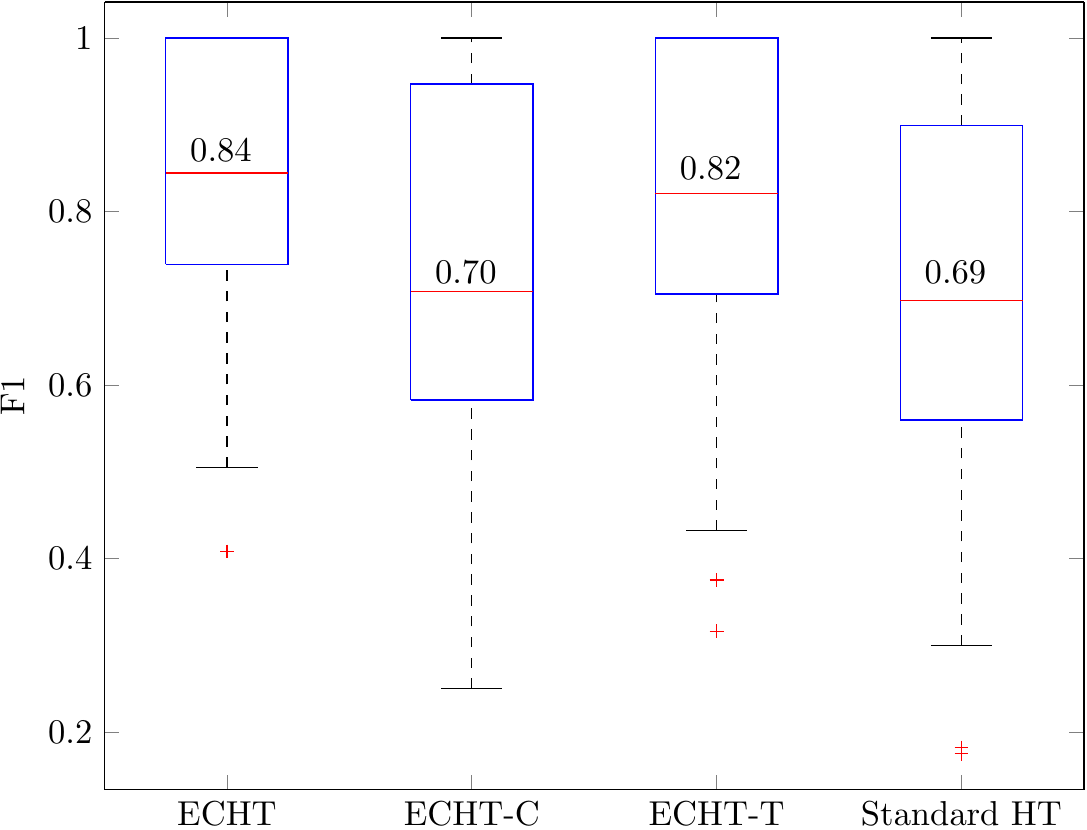}
	\caption{Hand action detection results for the variants of the proposed error correcting approach and the standard Hough transform method.}
	\label{fig:actDetectionRes}
\end{figure}

\begin{figure*}
	\centering\includegraphics[width=0.95\linewidth]{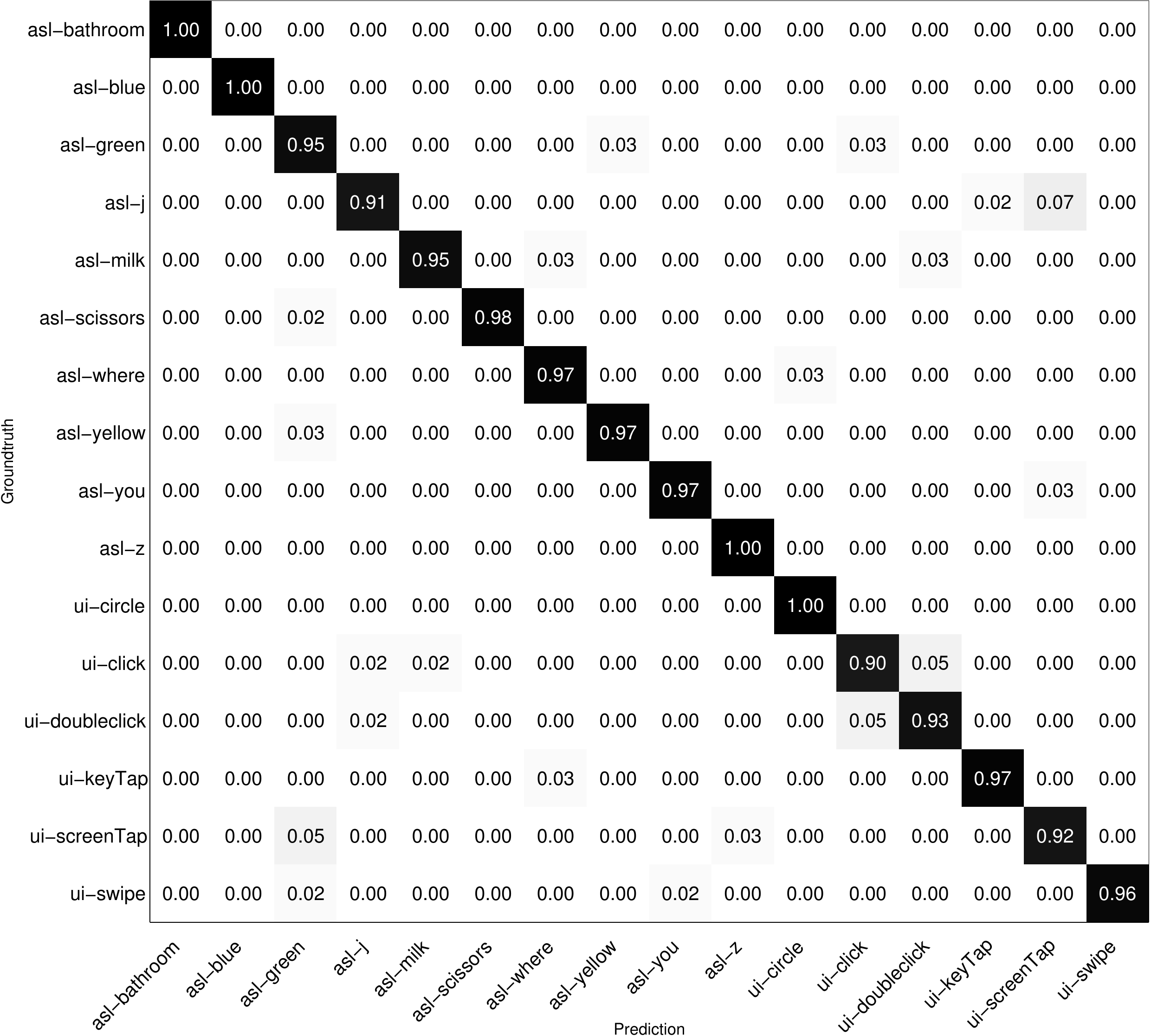}
	\caption{Confusion matrix of our ECHT on the action recognition task.}
	\label{fig:classification}
\end{figure*}

\subsection{Real-life Experiments}
Real-life experiments are conducted using our in-house dataset that contains ego-centric depth videos collected from 26 subjects of different ethnic groups and ages.
Some of the acquired hand data are with various accessories such as wrist watches, cords, rings, etc.
For the training data, we collect from 15 subjects 240 long video sequences which contain 2518 single-action subsequences.
For the testing data, we collect from another 11 subjects 60 long video sequences which contain 659 single-action subsequences.
The lengths of the long video sequences varying from 500 to 2500 frames,
while the length variation among the single-action subsequences from 7 frames up to 48 frames.

\paragraph{Action Recognition}
Action recognition task is based on single-action video clips. As stated previously, our in-house dataset contains 2518 training instances,
and 659 testing instances of 16 action classes.
The average F1 accuracy and standard deviation of the comparison methods are as follows:
variants of our approach ECHT \textbf{96.18\%$\pm$3.27\%}, ECHT-T 93.74\%$\pm$7.86\%, ECHT-C 88.40\%$\pm$18.74\%, the standard HT 87.89\%$\pm$18.66\%,
as well as the HMM1 and HMM2 of 63.64\%$\pm$23.57\% and 57.16\%$\pm$27.70\%, respectively.
Not surprisingly ECHT consistently overtakes the rest with a noticeable margin, which is followed by ECHT-T, ECHT-C, and standard HT, respectively,
while HMM1 \& HMM2 produce the least favorable results with much larger spread.
The confusion matrix of our \textbf{ECHT} is further presented in Fig.~\ref{fig:classification}.


%
%


\paragraph{Action Detection}
Our training data contains 240 long video sequences (with 2518 foreground single-action subsequences),
and the test data contains 60 long video sequences (with 659 foreground single-action subsequences) as mentioned previously,
These sequences also contain various background daily actions, such using keyboard \& mouse \& telephone, reading, writing, drinking,
looking around, etc.
Performances of the comparison methods are shown in Fig.~\ref{fig:actDetectionRes}, while Fig.~\ref{fig:visual} presents exemplar hand action detection visual results of our approach.
Experimental results suggest that temporal error correction is a primary factor accounting for the performance gains, while class error correction play a relatively minor role.
Moreover, although ECHT-C alone provides comparably small improvement,
The combined usage of temporal and class error correction always leads to notable performance gain over the baseline method of standard HT.
Moreover, our ECHT action detector is demonstrated being capable of robustly detecting the target actions from backgrounds.
Fig.~\ref{fig:falseDetExmples} shows several incorrect detection results, while more visual results are provided in the supplementary video files.

%

\begin{figure}
	\centering
	\includegraphics[width=0.44\columnwidth]{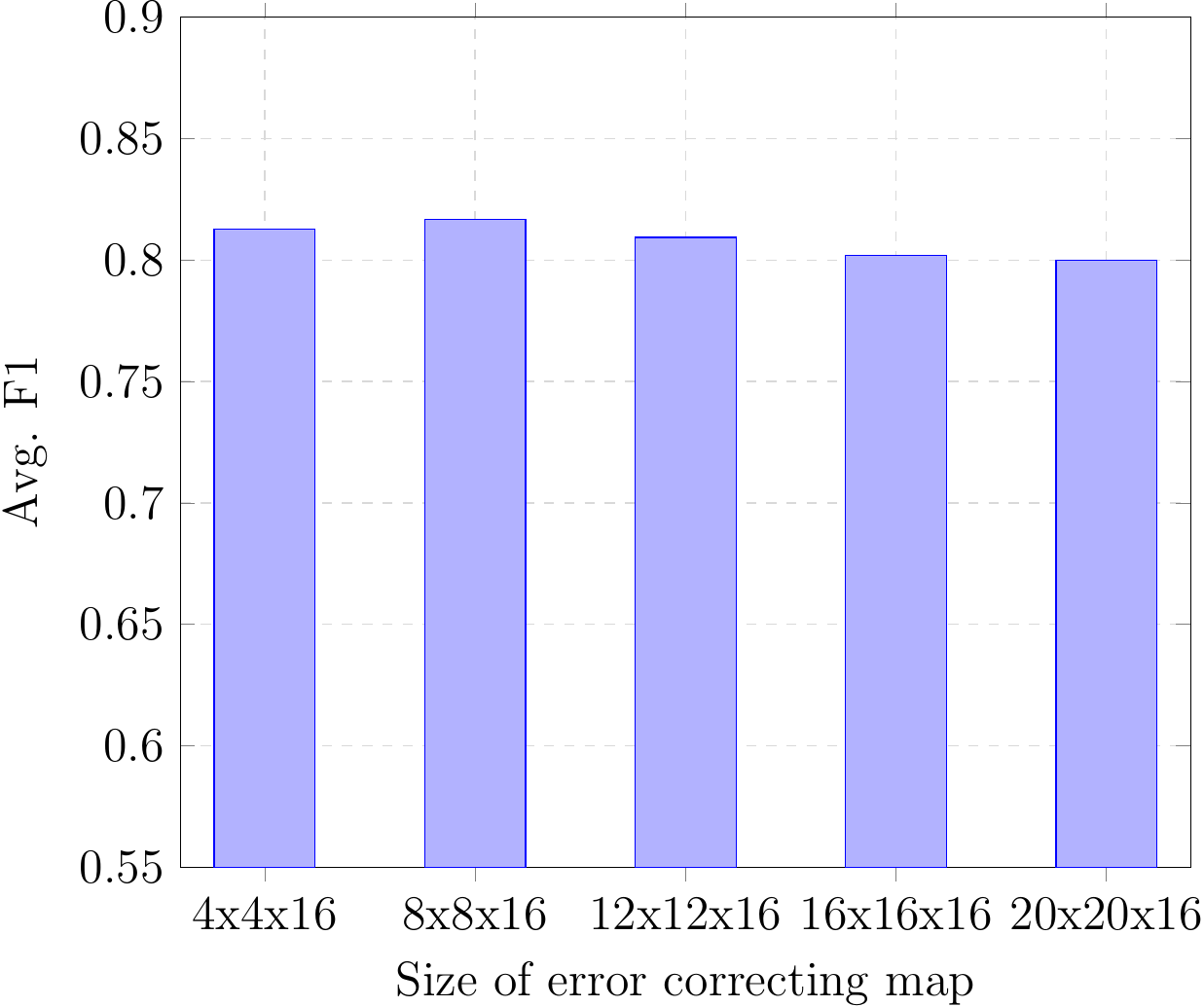}
\hspace{2mm}
	\includegraphics[width=0.44\columnwidth]{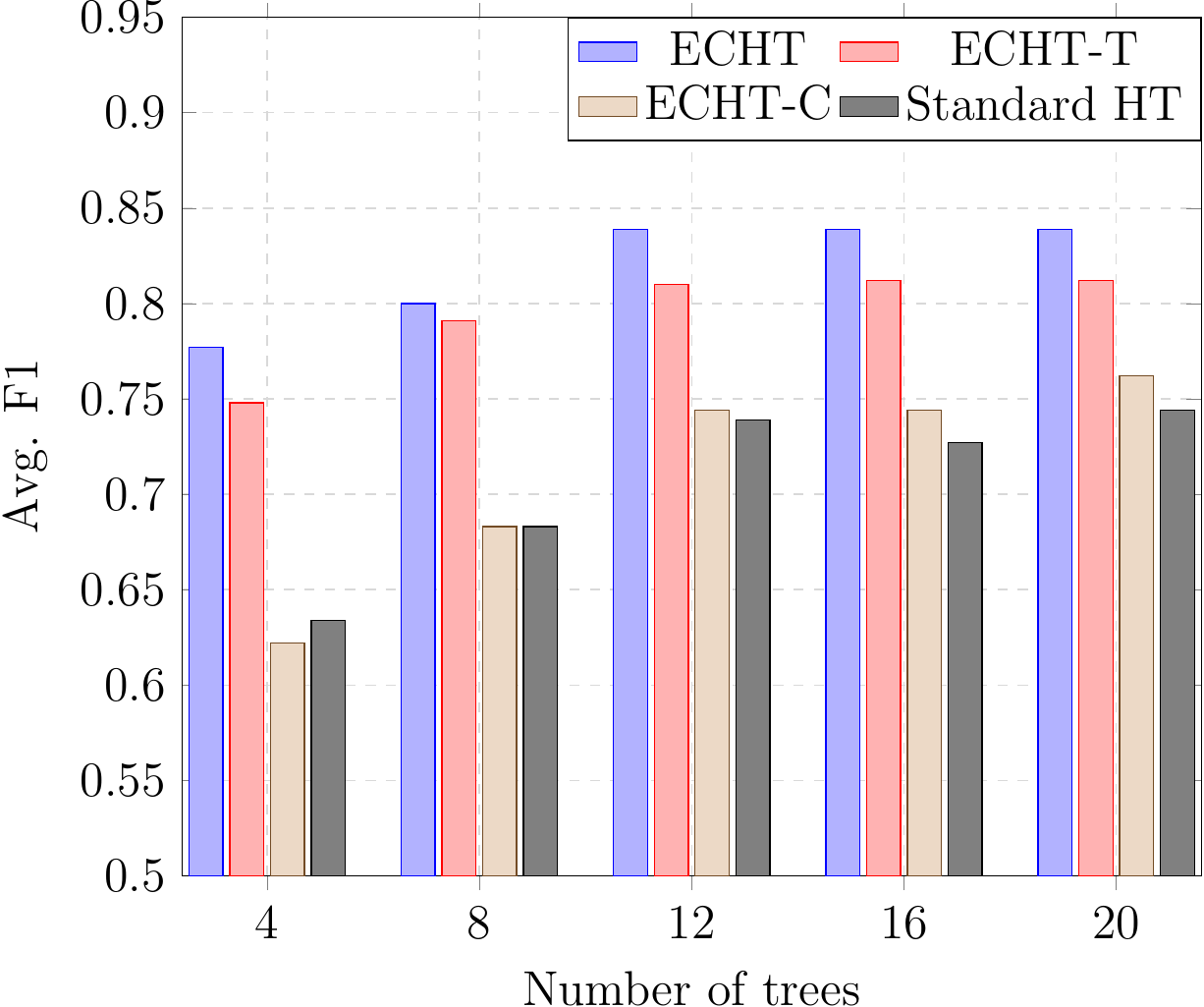}
	\caption{Robustness evaluation of the internal parameters. (a) F1 score as a function of the size of the error correcting map.
			(b) F1 score as a function of the number of trees, the same value is used for both $\mathcal{T}_c$ and $\mathcal{T}_{r}$. }
	\label{fig:internalParam}
\end{figure}

\paragraph{Size of error correcting map}
As our approach contain several internal parameters, it is of interest to examine the performance robustness with respect to these parameters.
We start by looking at the error correcting space, which is in our context a cubic space of $\delta \Lambda$,
$\delta \hat{\Lambda}$ and $\hat{y}$ that are quantized into $8 \times 8 \times 16$ sub-cubes.
The quantization on $\hat{y}$ is related to the number of classes.
We test the number of sub-cubes on $\delta \Lambda$ and $\delta \hat{\Lambda}$ from $4 \times 4$ to $20 \times 20$.
As in Fig.~\ref{fig:internalParam}(a), the hand action detection performance with respect to the size of the error correcting map is very stable.


\begin{figure}
	\centering
	\includegraphics[width=0.44\columnwidth]{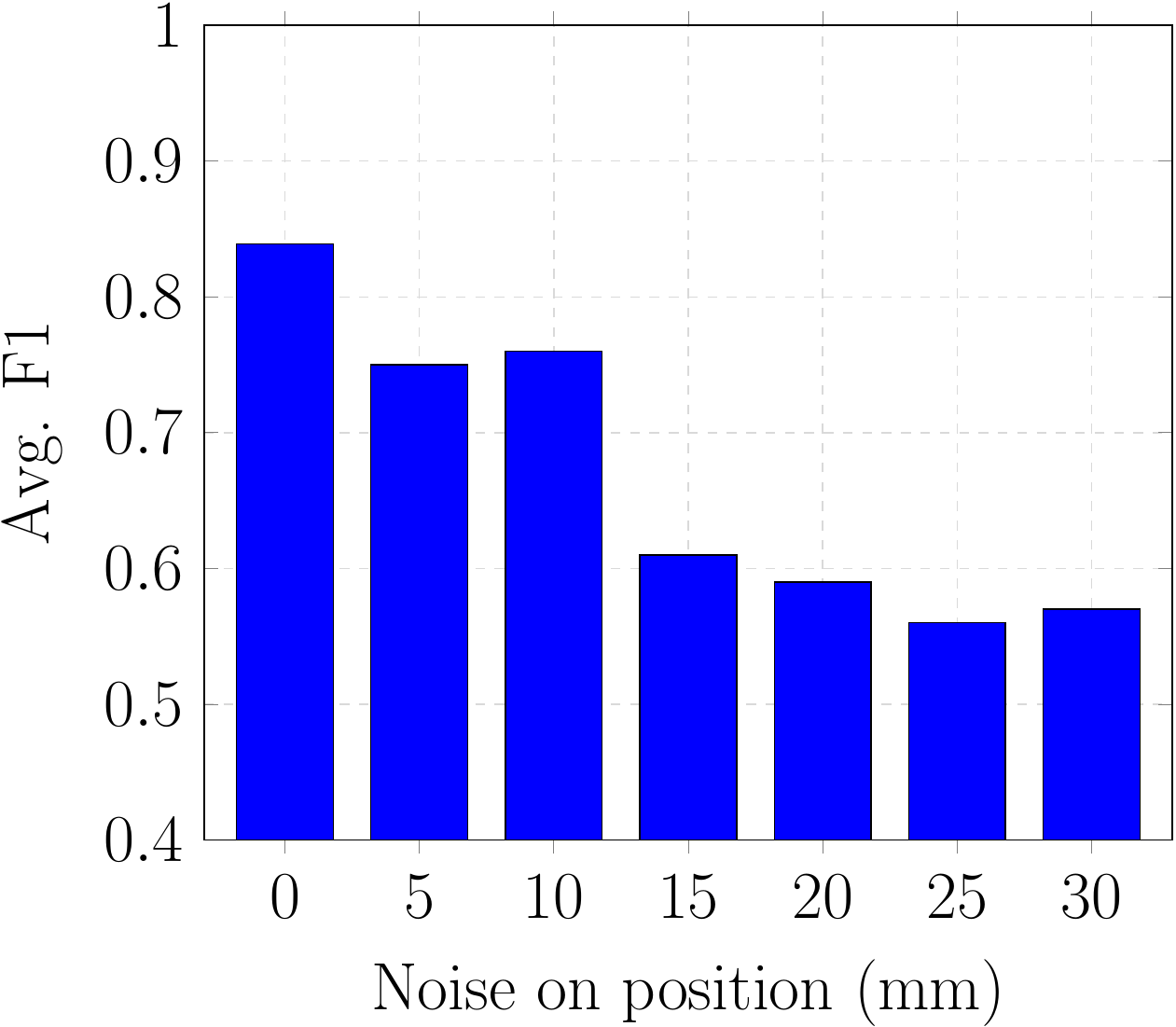}
\hspace{2mm}
	\includegraphics[width=0.44\columnwidth]{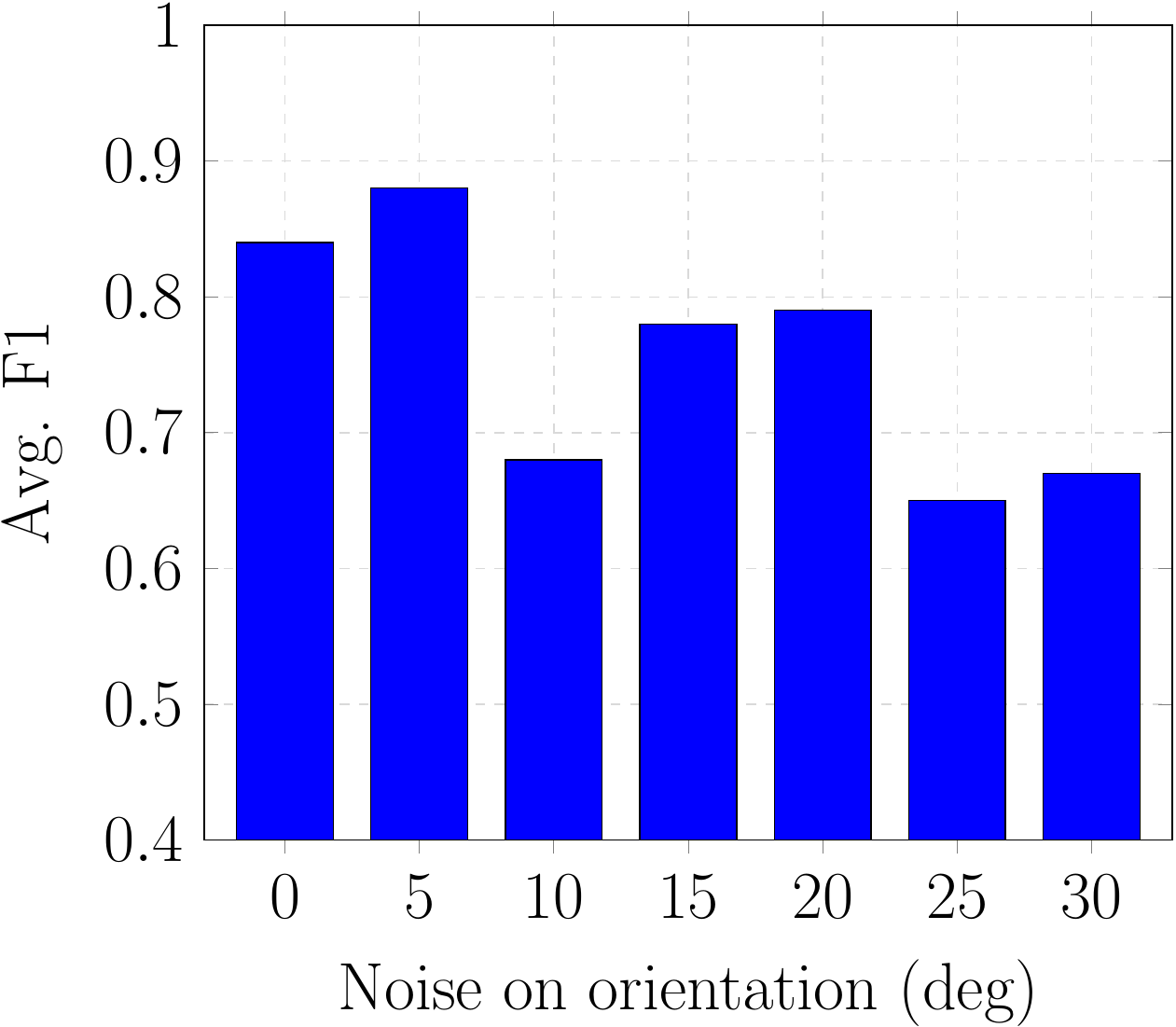}
%
	\caption{F1 score as a function of synthesized perturbations on hand location results.
			(a) F1 scores vs. the standard deviation of the Gaussian perturbation noise on hand locations.
			(b) F1 scores vs. the standard deviation of the Gaussian perturbation noise on hand orientations. }
	\label{fig:noise}
\end{figure}

\paragraph{Number of Trees}
We further investigate the performance variations with respect to the tree sizes in both $\mathcal{T}_c$ and $\mathcal{T}_{r}$.
As displayed in Fig.~\ref{fig:internalParam}(b), empirically the F1 score increases with the number of tree grows to 12,
then the F1 scores remain largely unchanged despite the continuing increased number of trees.


\paragraph{Perturbations of Hand Localization Results}
As our approach includes a preprocessing step to localize hand position and orientation, the result of this step will inevitably influences the overall performance.
To study its effect, we add random Gaussian noise as disturbance to the estimated hand location.
As displayed in Fig.~\ref{fig:noise}, when the perturbation noise on position is lower than 10~mm, the performance (F1 score) remains relatively stable.
When the noise reaches 15~mm and beyond, there starts to have a very noticeable performance drop.
This is mainly due to the fact that the estimated hand position frequently falls outside the hand region onto the background when noise is larger than 15~mm.
Nevertheless, in practice our hand location estimator can reliably locate the target in the hand region.
It is also observed that empirically our approach is relatively more stable with respect to the disturbance of orientation when comparing to that the disturbance of positions.

\paragraph{Running Time}

Real-life experiments are performed on a desktop with Intel Core i7 CPU 3.20GHz and 24GB memory. Note that our code is not optimized,
with only 1 core being used during the experiments.
Table~\ref{tab:efficiency} shows the result of time profiling on our algorithm for a test video sequence.
%


\begin{table}
	\centering
		\begin{tabular}{ | l | c | }
			\hline
			& Computation Time \\ \hline
			\hline
			Preprocess & 8.642 ms/frame \\ \hline
			Obtain Votes & 0.289 ms/frame \\ \hline
			Apply ECHT & 0.016 ms/frame \\ \hline
			\textbf{Total} & \textbf{8.947 ms/frame} \\ \hline
		\end{tabular}
	\caption{Computation time of an exemplar test run.}
	\label{tab:efficiency}
\end{table}

\section{Conclusion and Outlook}

This paper describes an error correcting Hough forest approach to tackle the novel and challenging problem of hand action detection from mobile ego-centric depth sequence.
Empirical evaluations demonstrate the applicability of the proposed approach. For future work, we plan to extend our approach to address action detection scenarios beyond hand actions.

{\small
\bibliographystyle{IEEEtran}
	\bibliography{main}

\begin{thebibliography}{10}
\providecommand{\url}[1]{#1}
\csname url@samestyle\endcsname
\providecommand{\newblock}{\relax}
\providecommand{\bibinfo}[2]{#2}
\providecommand{\BIBentrySTDinterwordspacing}{\spaceskip=0pt\relax}
\providecommand{\BIBentryALTinterwordstretchfactor}{4}
\providecommand{\BIBentryALTinterwordspacing}{\spaceskip=\fontdimen2\font plus
\BIBentryALTinterwordstretchfactor\fontdimen3\font minus
  \fontdimen4\font\relax}
\providecommand{\BIBforeignlanguage}[2]{{%
\expandafter\ifx\csname l@#1\endcsname\relax
\typeout{** WARNING: IEEEtran.bst: No hyphenation pattern has been}%
\typeout{** loaded for the language `#1'. Using the pattern for}%
\typeout{** the default language instead.}%
\else
\language=\csname l@#1\endcsname
\fi
#2}}
\providecommand{\BIBdecl}{\relax}
\BIBdecl

\bibitem{googleglass15}
``{G}oogle {G}lass,'' www.google.com/glass, 2013.

\bibitem{metaviewVR15}
``{M}etaview {S}paceglasses,'' www.getameta.com, 2015.

\bibitem{oculusVR15}
``{O}culus {R}ift,'' www.oculus.com, 2016.

\bibitem{FatFarReh:iccv11}
A.~Fathi, A.~Farhadi, and J.~Rehg, ``Understanding egocentric activities,'' in
  \emph{ICCV}, 2011.

\bibitem{LiFatReh:iccv13}
Y.~Li, A.~Fathi, and J.~Rehg, ``Learning to predict gaze in egocentric video,''
  in \emph{ICCV}, 2013.

\bibitem{HuaEtAl:cvpr15}
D.-A. Huang, M.~Ma, W.~Ma, and K.~Kitani, ``How do we use our hands?
  discovering a diverse set of common grasps,'' in \emph{CVPR}, 2015.

\bibitem{kinect11}
``Kinect,'' www.xbox.com/en-US/kinect, 2011.

\bibitem{softkinetic13}
``Softkinetic,'' www.softkinetic.com, 2012.

\bibitem{LiKit:cvpr13}
C.~Li and K.~M. Kitani, ``Pixel-level hand detection in ego-centric videos,''
  in \emph{CVPR}, 2013.

\bibitem{OikLouArg:cvpr14}
I.~Oikonomidis, M.~Lourakis, and A.~Argyros, ``Evolutionary quasi-random search
  for hand articulations tracking,'' in \emph{CVPR}, 2014.

\bibitem{TanEtAl:cvpr14}
D.~Tang, A.~Tejani, H.~Chang, and T.~Kim, ``Latent regression forest:
  Structured estimation of 3{D} articulated hand posture,'' in \emph{CVPR},
  2014.

\bibitem{QiaEtAl:cvpr14}
C.~Qian, X.~Sun, Y.~Wei, X.~Tang, and J.~Sun, ``Realtime and robust hand
  tracking from depth,'' in \emph{CVPR}, 2014.

\bibitem{XuEtAl:ijcv15}
C.~Xu, A.~Nanjappa, X.~Zhang, and L.~Cheng, ``Estimate hand poses efficiently
  from single depth images,'' \emph{IJCV}, pp. 1--25, 2015.

\bibitem{RegSupRam:cvpr15}
G.~Rogez, J.~Supancic, and D.~Ramanan, ``First-person pose recognition using
  egocentric workspaces,'' in \emph{CVPR}, 2015.

\bibitem{RegSupRam:iccv15}
------, ``Understanding everyday hands in action from rgb-d images,'' in
  \emph{ICCV}, 2015.

\bibitem{RazEtAl:eccv12}
N.~Razavi, J.~Gall, P.~Kohli, and L.~van Gool, ``latent {H}ough transform for
  object detection,'' \emph{ECCV}, 2012.

\bibitem{WohEtAl:bmvc12}
P.~Wohlhart, S.~Schulter, M.~Kostinger, P.~Roth, and H.~Bischof,
  ``discriminative {H}ough forests for object detection,'' \emph{BMVC}, 2012.

\bibitem{WooEtAl:ijcv13}
O.~Woodford, M.~Pham, A.~Maki, F.~Perbet, and B.~Stenger, ``Demisting the
  {H}ough transform for {3D} shape recognition and registration,'' \emph{IJCV},
  2013.

\bibitem{SchLapCap:icpr04}
C.~Schuldt, I.~Laptev, and B.~Caputo, ``Recognizing human actions: A local
  {SVM} approach,'' in \emph{ICPR}, 2004.

\bibitem{PirRam:cvpr14}
H.~Pirsiavash and D.~Ramanan, ``Parsing videos of actions with segmental
  grammars,'' in \emph{CVPR}, 2014.

\bibitem{SchGoo:cvpr08}
K.~Schindler and L.~van Gool, ``Action snippets: how many frames does human
  action recognition require?'' in \emph{CVPR}, 2008.

\bibitem{LapPer:iccv07}
I.~Laptev and P.~Perez, ``Retrieving actions in movies,'' in \emph{ICCV}, 2007.

\bibitem{YuaLiuWu:cvpr09}
J.~Yuan, Z.~Liu, and Y.~Wu, ``Discriminative {3D} subvolume search for
  efficient action detection,'' \emph{CVPR}, 2009.

\bibitem{GeoRosJit:iccv15}
G.~Gkioxari, R.~Girshick, and J.~Malik, ``Contextual action recognition with
  {R*CNN},'' in \emph{ICCV}, 2015.

\bibitem{YaoGalGoo:cvpr10}
A.~Yao, J.~Gall, and L.~V. Gool, ``A {H}ough transform-based voting framework
  for action recognition,'' in \emph{CVPR}, 2010.

\bibitem{YuKimCip:cvpr13}
T.~Yu, T.~Kim, and R.~Cipolla, ``Unconstrained monocular 3d human pose
  estimation by action detection and cross-modality regression forest,
  booktitle={CVPR}, year={2013},.''

\bibitem{PirRam:cvpr12}
H.~Pirsiavash and D.~Ramanan, ``Detecting activities of daily living in
  first-person camera views,'' in \emph{CVPR}, 2012.

\bibitem{RyoMat:cvpr13}
M.~Ryoo and L.~Matthies, ``First-person activity recognition: What are they
  doing to me?'' in \emph{CVPR}, 2013.

\bibitem{DuWanWan:cvpr15}
Y.~Du, W.~Wang, and L.~Wang, ``Hierarchical recurrent neural network for
  skeleton based action recognition,'' in \emph{CVPR}, 2015.

\bibitem{DonEtAl:cvpr15}
J.~Donahue, L.~Hendricks, S.~Guadarrama, M.~Rohrbach, S.~Venugopalan,
  K.~Saenko, and T.~Darrell, ``Long-term recurrent convolutional networks for
  visual recognition and description,'' in \emph{CVPR}, 2015.

\bibitem{YeEtAl:bookchapter13}
M.~Ye, Q.~Zhang, L.~Wang, J.~Zhu, R.~Yang, and J.~Gall, \emph{Time-of-Flight
  and Depth Imaging: Sensors, Algorithms, and Applications}, ser. LNCS
  8200.\hskip 1em plus 0.5em minus 0.4em\relax Springer, 2013, ch. A Survey on
  Human Motion Analysis from Depth Data, pp. 149--87.

\bibitem{WanEtAl:pami14}
J.~Wang, Z.~Liu, Y.~Wu, and J.~Yuan, ``Learning actionlet ensemble for 3{D}
  human action recognition,'' \emph{IEEE TPAMI}, vol.~36, no.~5, pp. 914--27,
  2014.

\bibitem{WeiEtAl:iccv13}
P.~Wei, N.~Zheng, Y.~Zhao, and S.~Zhu, ``Concurrent action detection with
  structural prediction,'' in \emph{ICCV}, 2013.

\bibitem{MogEtAl:cvprwshp14}
M.~Moghimi, P.~Azagra, L.~Montesano, A.~Murillo, and S.~Belongie, ``Experiments
  on an {RGB-D} wearable vision system for egocentric activity recognition,''
  in \emph{CVPR Workshop on Egocentric (First-person) Vision}, 2014.

\bibitem{LeeKim:tmapi99}
H.-K. Lee and J.~H. Kim, ``An {HMM}-based threshold model approach for gesture
  recognition,'' \emph{IEEE TPAMI}, vol.~21, no.~10, pp. 961--73, 1999.

\bibitem{ShoEtAl:cacm13}
J.~Shotton, T.~Sharp, A.~Kipman, A.~Fitzgibbon, M.~Finocchio, A.~Blake,
  M.~Cook, and R.~Moore, ``Real-time human pose recognition in parts from
  single depth images,'' \emph{Comm. ACM}, vol.~56, no.~1, pp. 116--24, 2013.

\bibitem{lang2012sign}
S.~Lang, M.~Block, and R.~Rojas, ``Sign language recognition using kinect,'' in
  \emph{Artificial Intelligence and Soft Computing}.\hskip 1em plus 0.5em minus
  0.4em\relax Springer, 2012, pp. 394--402.

\bibitem{molchanov2015multi}
P.~Molchanov, S.~Gupta, K.~Kim, and K.~Pulli, ``Multi-sensor system for
  driver’s hand-gesture recognition,'' in \emph{IEEE Conference on Automatic
  Face and Gesture Recognition}, 2015.

\bibitem{ohn2014hand}
E.~Ohn-Bar and M.~M. Trivedi, ``Hand gesture recognition in real time for
  automotive interfaces: A multimodal vision-based approach and evaluations,''
  \emph{IEEE Trans on Intelligent Transportation Systems}, vol.~15, no.~6, pp.
  2368--77, 2014.

\bibitem{Hou:ICHEAI59}
P.~Hough, ``Machine analysis of bubble chamber pictures,'' in \emph{Proc. Int.
  Conf. High Energy Accelerators and Instrumentation}, 1959.

\bibitem{DudHar:CACM72}
R.~Duda and P.~Hart, ``Use of the {H}ough transformation to detect lines and
  curves in pictures,'' \emph{Commun. ACM}, vol.~15, pp. 11--5, 1972.

\bibitem{Bal:PR81}
D.~Ballard, ``Generalizing the {H}ough transform to detect arbitrary shapes,''
  \emph{Pattern Recognition}, vol.~13, no.~2, pp. 111--22, 1981.

\bibitem{LeiLeoSch:eccv04whp}
B.~Leibe, A.~Leonardis, and B.~Schiele, ``Combined object categorization and
  segmentation with an implicit shape model,'' in \emph{ECCV Workshop
  statistical learning in CV}, 2004.

\bibitem{MajMal:cvpr09}
S.~Maji and J.~Malik, ``Object detection using a max-margin {H}ough
  transform,'' in \emph{CVPR}, 2009.

\bibitem{GalEtAl:pami11}
J.~Gall, A.~Yao, N.~Razavi, L.~V. Gool, and V.~Lempitsky, ``{H}ough forests for
  object detection, tracking, and action recognition,'' \emph{IEEE TPAMI},
  vol.~33, no.~11, pp. 2188--202, 2011.

\bibitem{yarlagadda2010voting}
P.~Yarlagadda, A.~Monroy, and B.~Ommer, ``Voting by grouping dependent parts,''
  in \emph{ECCV}, 2010.

\bibitem{barinova2012detection}
O.~Barinova, V.~Lempitsky, and P.~Kholi, ``On detection of multiple object
  instances using {H}ough transforms,'' \emph{IEEE TPAMI}, vol.~34, no.~9, pp.
  1773--84, 2012.

\bibitem{HufPle:book03}
W.~Huffman and V.~Pless, \emph{Fundamentals of error-correcting codes}.\hskip
  1em plus 0.5em minus 0.4em\relax Cambridge Univ. Press, 2003.

\bibitem{DieBak:jair95}
T.~Dietterich and G.~Bakiri, ``Solving multiclass learning problems via
  error-correcting output codes,'' \emph{JAIR}, vol.~2, pp. 263--86, 1995.

\bibitem{shreyas2009}
S.~B. Guruprasad, ``On the breakdown point of the {H}ough transform,'' in
  \emph{International Conference on Advances in Pattern Recognition}, 2009.

\bibitem{ChaLin:atist11}
C.~Chang and C.~Lin, ``{LIBSVM}: A library for support vector machines,''
  \emph{ACM Trans. Intel. Sys. Tech.}, vol.~2, pp. 1--27, 2011.

\end{thebibliography}
}

\end{document}